\pdfoutput=1

\documentclass[11pt]{article}

\usepackage[preprint]{acl}

\usepackage{times}
\usepackage{latexsym}

\usepackage[T1]{fontenc}

\usepackage[utf8]{inputenc}
\usepackage{amsfonts}
\usepackage{amsmath}
\usepackage{amssymb}
\usepackage{amsthm}
\usepackage{enumitem}
\usepackage{listings}
\definecolor{darkpink}{rgb}{0.96,0.14,0.56}
\usepackage{hyperref} 
 \hypersetup{ 
     colorlinks=true, 
     linkcolor=blue, 
     filecolor=blue, 
     urlcolor=darkpink, 
 } 
\usepackage{bbm}
\DeclareMathOperator*{\argmax}{arg\,max}

\usepackage{microtype}

\usepackage{inconsolata}

\usepackage{graphicx}

\title{\fontsize{15}{16}\selectfont Encoding and Controlling Global Semantics for Long-form Video Question Answering}

\author{
  Thong Thanh Nguyen$^{1}$\thanks{Corresponding to: Thong, \href{e0998147@u.nus.edu}{e0998147@u.nus.edu}} \quad Zhiyuan Hu$^{1}$ \quad Xiaobao Wu$^{2}$ \quad Cong-Duy T Nguyen$^{2}$ \\
  \quad \textbf{See-Kiong Ng}$^{1}$ \quad \textbf{Anh Tuan Luu}$^{2}$ 
  \\
  $^1$ National University of Singapore \quad
  $^2$ Nanyang Technological University }

\begin{document}
\maketitle
\begin{abstract}

Seeking answers effectively for long videos is essential to build video question answering (videoQA) systems.
Previous methods adaptively select frames and regions from long videos to save computations. However, this fails to reason over the whole sequence of video, leading to sub-optimal performance.
To address this problem, we introduce a state space layer (SSL) into multi-modal Transformer to efficiently integrate global semantics of the video, which mitigates the video information loss caused by frame and region selection modules. Our SSL includes a gating unit to enable controllability over the flow of global semantics into visual representations. To further enhance the controllability, we introduce a cross-modal compositional congruence (C$^3$) objective to encourage global semantics aligned with the question. To rigorously evaluate long-form videoQA capacity, we construct two new benchmarks Ego-QA and MAD-QA featuring videos of considerably long length, \textit{i.e.} 17.5 minutes and 1.9 hours, respectively. Extensive experiments demonstrate the superiority of our framework on these new as well as existing datasets. The code, model, and data have been made available at \href{https://nguyentthong.github.io/Long_form_VideoQA}{nguyentthong.github.io/Long\_form\_VideoQA}.
\end{abstract}

\section{Introduction}

VideoQA has been extensively studied to develop systems to assist humans in daily activities \citep{grauman2022ego4d, lei2021assistsr}, \textit{e.g.}, remind users of their past actions, help users locate their belongings, and provide assistance with complex tasks. To implement these functions, we expect videoQA systems  to understand and extract relevant information from long-form videos with diverse objects and complex spatial-temporal interactions.

\begin{figure*}[t]
    \centering
    \includegraphics[width=\linewidth]{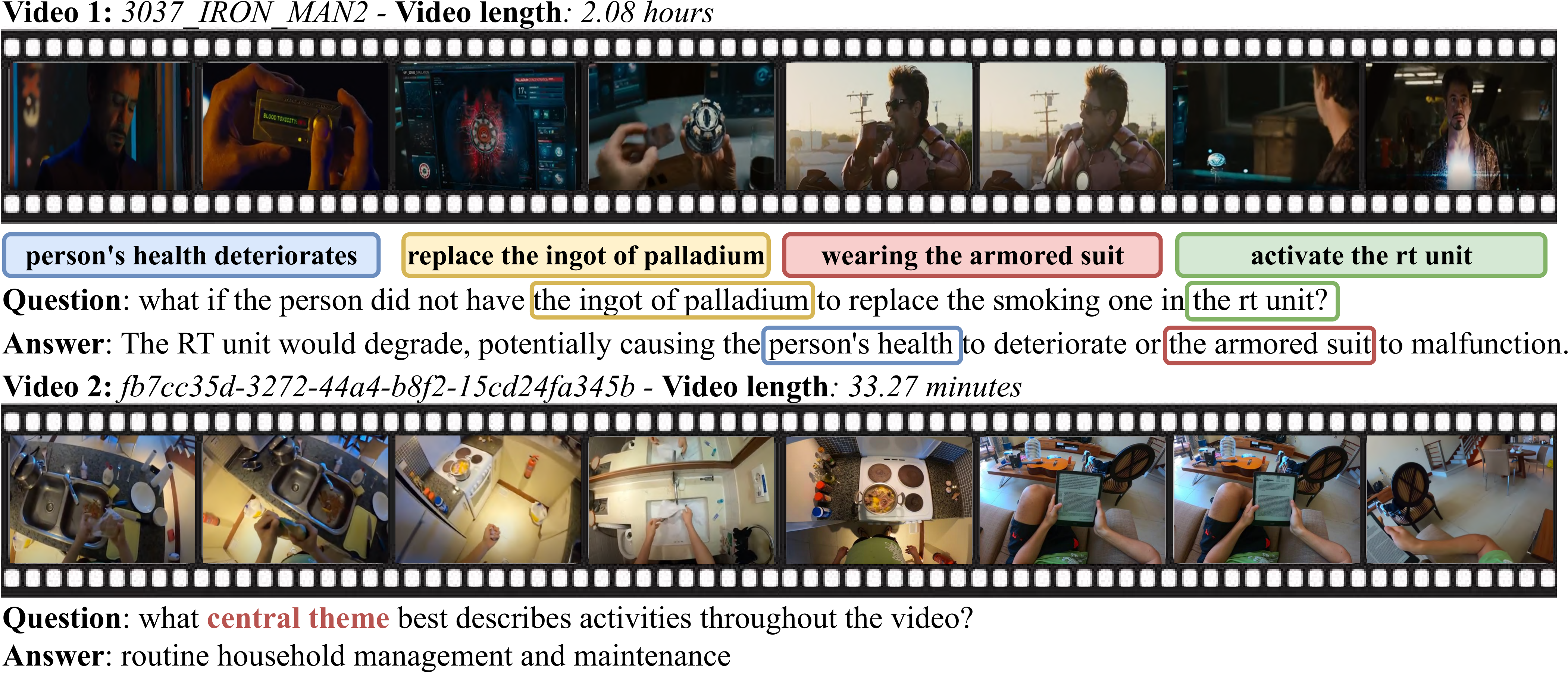}
    \caption{Long-form videoQA examples, with videos taken from MAD \citep{soldan2022mad} and Ego4D \citep{grauman2022ego4d} datasets, respectively. Question in video 1 requires the model to reason about the relation chain of replacing \textit{ingot of palladium} to activate the \textit{rt unit} that powers the \textit{armored suit} and protects \textit{person’s health}. Question in video 2 necessitates an understanding of the overall theme in video 2.}
    \label{fig:longform_videoqa_example}
\end{figure*}

Compared with short clips, long-form videos pose more challenges for videoQA.
They consist of a higher number of objects and events. As such, comprehensively encoding information from them requires expensive computations. Moreover, a high amount of information may be unrelated to the posed question. To address these problems, recent studies \citep{bain2021frozen, wang2023all, gao2023mist} adaptively select a subset of video frames and visual regions associated with the question. Nevertheless, if a question necessitates a reasoning of the entire sequence of events (\textit{e.g.} video 1’s Figure \ref{fig:longform_videoqa_example}), or an understanding of the overall video narration (\textit{e.g.} video 2’s Figure \ref{fig:longform_videoqa_example}), a mere handful of selected frames or regions might not sufficiently encapsulate necessary details.

To tackle these problems, we introduce a state space layer (SSL). Before forwarding video frames to selection modules, SSL fixes long-term dependency patterns for integrating global information into visual representations. Such global information offers the selected frames the global context within the video, so that they can relate to other frames even though those frames are not selected for attention computation. However, a considerable amount of unrelated global information may flow into visual representations. Therefore, we first equip SSL with a gating mechanism to provide more controllability over the flow of global semantics into visual representations, resulting in our \textbf{G}ated \textbf{S}tate space \textbf{M}ulti-modal \textbf{T}ransformer (GSMT) architecture. 
Furthermore, we promote global semantics that is more aligned with the question. In particular, we introduce \textbf{C}ross-modal \textbf{C}ompositional \textbf{C}ongruence (C$^3$) objective that compares visual attention with its version transitioned to the language basis via cross-modal attention, effectively measuring cross-modal congruence between intra-modal relations. Our rationale behind focusing on intra-modal relations is because videoQA models often need to understand spatial and temporal relationships between entities and events posed by the question \citep{gandhi2022measuring}, thus we encourage globally informed visual representations to maintain compositional consistency between visual patches and question entities.

Remarkably, we observe that recent long-form videoQA works \citep{gao2023mist, islam2024video} still mostly evaluate on videos lasting at most one minute or two, and use short-natured questions which necessitate watching only a short period of video, \textit{i.e.}, about 100 seconds  \citep{mangalam2023egoschema}, to determine the answer. To more rigorously evaluate long-form videoQA capacity, we introduce a construction procedure which utilizes large language model (LLM) to generate questions and associated answers for egocentric and movie videos whose average lengths are 17.5 minutes and 1.9 hours, respectively. Additionally, we also conduct automatic and manual filtering to obtain high-quality questions which require watching a video up to 1200 seconds to answer, longer than any existing long-form videoQA benchmarks \citep{xiao2021next, wu2021star, mangalam2023egoschema}.

To sum up, our contributions are as follows:
\begin{itemize}[leftmargin=*]
    \item We propose a Gated State space Multi-modal Transformer (GSMT) with state space layer (SSL) to integrate global information into visual representations for long-form videoQA.
    
    \item We equip SSL with a gating mechanism to provide controllability over the flow of global video semantics and a Cross-modal Compositional Congruence (C$^3$) objective to encourage question-aligned visual representations. 
    
    \item We curate two new datasets with excessively long video lengths and long-natured questions for long-form videoQA. Comprehensive experiments on our curated and five standard datasets substantiate the superiority of our framework over various competitive baselines.
\end{itemize}

\section{Related Work}
\noindent\textbf{Video question answering (videoQA).} VideoQA datasets \citep{xu2017video, jang2017tgif, tapaswi2016movieqa, lei2018tvqa, nguyen2024video} mostly focus on short video clips about daily activities, \textit{e.g.} sports or household work. Recent works \citep{gao2021env, grunde2021agqa, wu2021star} focus on complex spatial-temporal reasoning ability over longer temporal lengths with causal and transition questions. Regarding methodology, early works \citep{zhao2018open, li2019beyond} present LSTM-based and GNN-based architectures to capture cross-modal and motion-appearance interaction.  For example, \citep{xiao2023contrastive} incorporate graph modeling into Transformer to explicitly encode the object relations in videos. Due to outstanding performance of pre-trained vision-language Transformers, \citet{bain2021frozen, fu2023empirical, wang2023all} utilize pre-trained Transformer models on downstream videoQA tasks. 

\noindent\textbf{Long-form video modeling.} To improve the applicability of vision-language systems, various works have focused on long-form video modeling for videoQA tasks.  \citet{wu2021towards} introduce short-term feature extraction technique and long-term memory mechanism to alleviate redundant video frame processing. \citet{lin2022eclipse} propose to compensate sparsely extracted video frames with audio cues. \citet{islam2022long} design structured multi-scale temporal decoder. \citet{gao2023mist} utilize question as a guide to select question-related visual segments to mitigate computation. Inspired by \citep{nguyen2021enriching, wu2024survey} in language modeling domains, we introduce a method to encode global semantics of videos to tackle information degradation problem in long-form video modeling.


\noindent\textbf{Cross-modal alignment.} Alignment approaches, such as contrastive learning \citep{nguyen2021contrastive, nguyen2024topic,wu2023infoctm,wu2024modeling}, optimal transport \citep{nguyen2022improving, wu2023effective,wu2024affinity, wu2024fastopic}, tree-based fusion \citep{nguyen2023gradient}, and energy-based modeling \citep{nguyen2023demaformer}, have been investigated to compare different input samples. For multimodal learning, recent cross-modal alignment methods focus on establishing a shared latent space in which samples of different modalities can be compared readily \citep{nguyen2023expand, nguyen2024meta, nguyen2024read, nguyen2024kdmcse, nguyen2023improving, nguyen2022adaptive, nguyen2022vision}. For example, with a contrastive learning formulation, CLIP \citep{radford2021learning} and ALIGN \citep{jia2021scaling} learn generalizable image-text representations from millions of image-text pairs. To exploit relation for cross-modal alignment, \citet{ren2021learning} use hard bijective correspondences between words and objects via an argmax over the cross-modal attention matrix to optimize the cross-modal alignment. \citet{pandey2022cross} generalize the hard alignment with their soft alignment approach, but they focus on objects within an image rather than a long video. 

\section{Methodology}
\begin{figure*}[t]
    \centering
    \includegraphics[width=\linewidth]{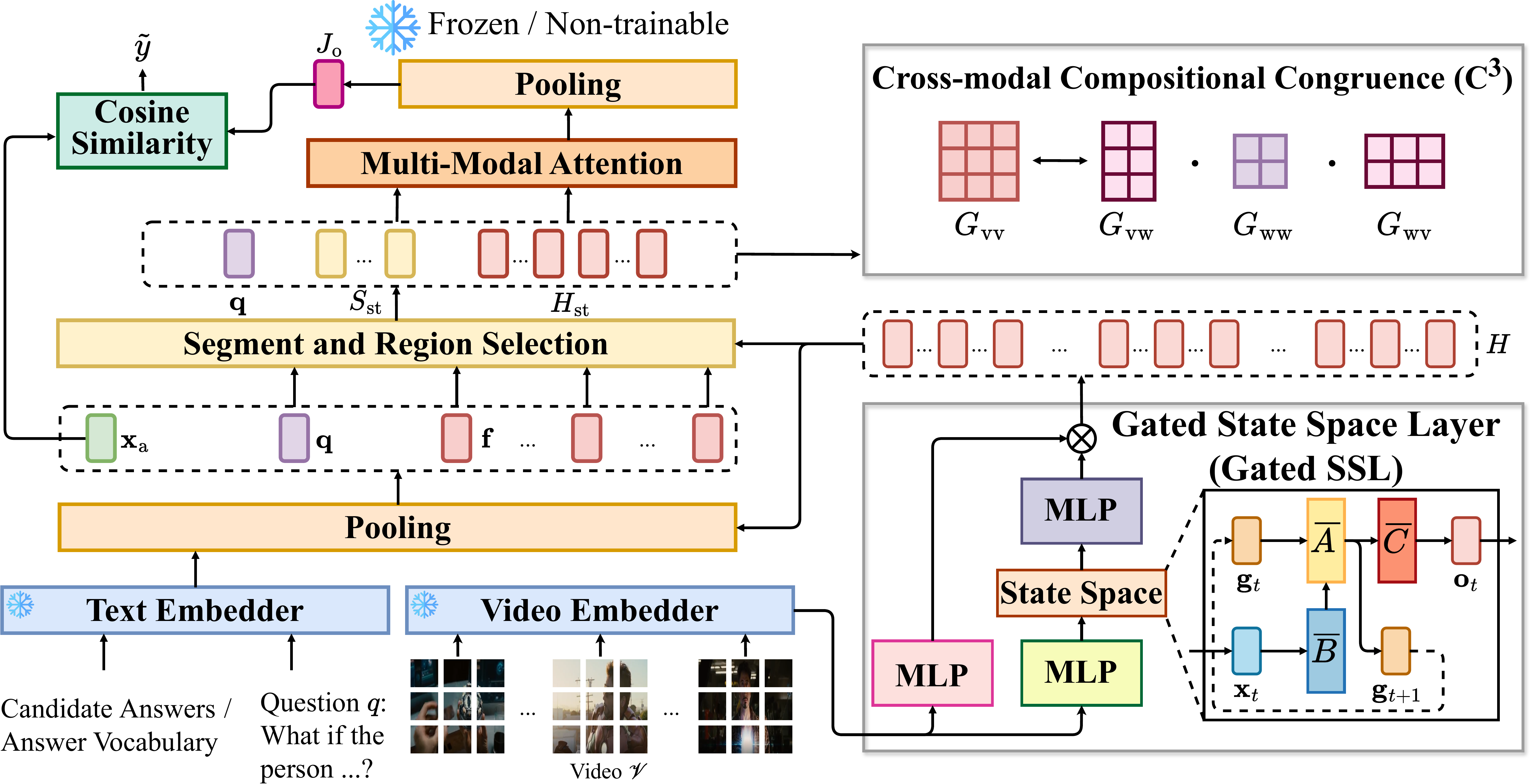}
    \caption{Illustration of the GSMT architecture empowered by gated SSL and C$^3$ training objective.}
    \label{fig:overall_architecture}
\end{figure*}

The formulation of video question answering (videoQA) is to predict the answer $y$ for a question $q$ about a video $\mathcal{V}$ as follows:
\begin{equation}
\tilde{y} = \argmax_{y \in \mathcal{A}} \mathcal{F}_{\theta} (y|q, \mathcal{V}, \mathcal{A}),
\end{equation}
where $\tilde{y}$ denotes the predicted answer chosen from the set of candidate options $\mathcal{A}$, and $\theta$ denotes the trainable parameters of a videoQA model. With this videoQA task formulation, we explain our proposed GSMT architecture as the model $\mathcal{F}_{\theta}$ and C$^3$ objective to support GSMT. The overall framework is illustrated in Figure \ref{fig:overall_architecture}.

\subsection{Gated State Space Multi-Modal Transformer (GSMT)}
Our GSMT takes videos and questions as input, divides each video frame into visual patches and a question into textual words, and then forwards visual patches and textual words into video and text embedder to extract initial representations.

\noindent\textbf{Input Embedder.} For video embedder, we utilize frozen pre-trained vision-language Transformer to extract patch-level features $X = \{\mathbf{x}_{0}, \mathbf{x}_{1}, ..., \mathbf{x}_{L-1}\}$, of all $T$ frames, where $t$-th frame consists of $N$ patch-level features $\{\mathbf{x}_{t,j}\}_{j=0}^{N-1}$, hence $L = NT$, and $\mathbf{x} \in \mathbb{R}^{d}$. For text embedder, a similar frozen pre-trained vision-language Transformer is used to obtain word-level features $W = \{\mathbf{w}_{0}, \mathbf{w}_{1}, ..., \mathbf{w}_{M-1}\}$, where $\mathbf{w}_{0}$ corresponds to the \texttt{[CLS]} token and $\mathbf{w}_{1}, ..., \mathbf{w}_{M-1}$ correspond to words in the question.

\noindent\textbf{Gated State Space Layer (Gated SSL).} Inspired by \citep{gu2021efficiently}, we define a sequence-to-sequence map $SSL(X)$ from a sequence of patch-level features to $d_{S}$-dim hidden states, parameterized by learnable state matrices $A \in \mathbb{R}^{d_{S} \times d}, B \in \mathbb{R}^{d_{S} \times d}, C \in \mathbb{R}^{d \times d_{S}}$, and step size $\Delta$ as:
\begin{gather}
\mathbf{g}_{t+1} = \overline{A} \cdot \mathbf{g}_{t} + \overline{B} \cdot \mathbf{x}_{t+1}, \quad \mathbf{o}_{t+1} = \overline{C} \cdot \mathbf{g}_{t+1}, \\
\overline{A} = e^{A\Delta}, \quad \overline{B} = (\overline{A} - I) A^{-1} B, \quad \overline{C} = C.
\end{gather}
We unroll the mapping to obtain:
\begin{equation}
\mathbf{o}_{t} = \sum\limits_{j=0}^{t} \overline{CA}^{j}\overline{B} \cdot \mathbf{x}_{t-j}.
\end{equation}
This can be written as a convolutional representation $O = \overline{\Gamma} * X$, where $\overline{\Gamma} = \left(\overline{CB}, \overline{CAB}, ..., \overline{CA}^{L-1}\overline{B}\right)$ denotes the convolutional kernel, $*$ the discrete convolution operator, $X$ the input sequence, $O$ the corresponding output sequence. This convolution denotes the fixed global dependency pattern that facilitates the computation of global information among visual patches. We use Fast Fourier Transformer (FFT) \citep{cooley1965algorithm} to compute the convolution in parallel provided that $\bar{\Gamma}$ has been obtained. 

Computing the kernel $\bar{\Gamma}$ is non-trivial since it requires $L$ distinct matrix powers. Instead, inspired by \citep{gupta2022diagonal}, we initialize $A$ to be a diagonal matrix $\text{diag}\left(\lambda_{1}, \lambda_{2}, ..., \lambda_{d_{S}}\right)$ and $B$ to all-one matrix $\mathbbm{1}_{d_{S} \times d}$. Due to this initialization, the kernel can be computed as:
\begin{equation}
\overline{\Gamma} = \left(C \odot E\right) \cdot P,
\end{equation}
where $E = \left(\frac{e^{\lambda_{1}\Delta-1}}{\lambda_{1}}, \frac{e^{\lambda_{2}\Delta-1}}{\lambda_{2}}, ..., \frac{e^{\lambda_{d_{S}}\Delta-1}}{\lambda_{d_{S}}}\right) \in \mathbb{R}^{d_{S}}$, in which $P_{i,j} = e^{\lambda_{i} \cdot j \cdot \Delta} \in \mathbb{R}^{d_{S} \times L}$, and $\odot$ denotes the element-wise multiplication.

To equip SSL with the control over which global semantics to integrate into visual representations, we construct a gating unit as: 
\begin{gather}
U = \phi\left(\text{Linear}(X)\right), \;\; V = \phi\left(\text{Linear}(X)\right), \\
O = \text{Linear}\left(\text{SSM}\left(U\right)\right), \\
H = \text{Linear}\left(O \odot V\right),
\end{gather}
where $U, V, O \in \mathbb{R}^{L \times d_{\text{gating}}}$ denote the intermediate gating representations, $H \in \mathbb{R}^{L \times d_{h}}$ the visual representations output by gated SSL, $\phi$ the non-linear activation, and $d_{\text{gating}} < d, d_{h}, d_{s}$.

\noindent\textbf{Visual Segment and Region Selection.} After obtaining visual patch-level hidden representations $H$, we proceed to obtain frame features by pooling each frame $t$’s visual patches:
\begin{equation}
\mathbf{f}_{t} = \text{pool}\left(\mathbf{h}_{t,0}, \mathbf{h}_{t,1}, ..., \mathbf{h}_{t,N-1} \right).
\end{equation}
Subsequently, we group non-overlapping consecutive frames into a segment to obtain $I$ segments, each of which contains $N_{p} = \lceil\frac{L}{I}\rceil$ patches and $N_{f} = \lceil \frac{T}{I}\rceil$ frames. We proceed to compute segment features through pooling frame features corresponding to the segment:
\begin{equation}
\mathbf{s}_{i} = \text{pool}\left(\mathbf{f}_{i,0}, \mathbf{f}_{i,1}, ..., \mathbf{f}_{i,N_{f}-1}\right).
\end{equation} 
Similarly, we also pool word features to obtain the question representation:
\begin{equation}
\mathbf{q} = \text{pool}\left(\mathbf{w}_{0}, \mathbf{w}_{1}, ..., \mathbf{w}_{M-1}\right).
\end{equation}
Given the segment features $S = \{\mathbf{s}_{i}\}_{i=0}^{I-1}$ and question feature $\mathbf{q}$, we conduct top-$k$ segment selection:
\begin{gather}
Q = \text{Linear}\left(\mathbf{q}\right),\; K = \text{Linear}\left(S\right), \\
\mathcal{B} = \mathop{selector}_{Top_{k}} \left(\text{softmax}\left(\frac{QK^{\top}}{\sqrt{d_{k}}}\right)\right).
\end{gather}
We implement $selector$ as a differentible Gumbel-softmax selection function. The output of the $selector$ is a sequence of segment index $\mathcal{B}$. Thereby, we extract respective segment features with respect to the selected segment indices, \textit{i.e.} $S_{\text{st}} = \{\mathbf{s}_{b} \;|\; b \in \mathcal{B}\}$.

For every frame $\tau$ in the selected segments, we then perform cross-modal attention between its patch-level hidden representations $H_{\tau} = \{\mathbf{h}_{\tau,j}\}_{j=0}^{N-1}$, $\tau \in \{\lfloor\frac{b}{N_{f}}\rfloor \; | \; b \in \mathcal{B}\}$ and the question to select top-$j$ question-related patches:
\begin{gather}
Q = \text{Linear}\left(\mathbf{q}\right), \; K = \text{Linear}\left(H_{\tau}\right), \\
\mathcal{T} = \mathop{selector}_{Top_{j}} \left(\text{softmax}\left(\frac{QK^{\top}}{\sqrt{d_{k}}}\right)\right).
\end{gather}
Lastly, we stack the selected patches of all selected frames to obtain $H_{\text{st}} = \{\mathbf{h}_{\tau,j}|\tau \in \mathcal{T}\}_{j=0}^{N-1}$.

\noindent\textbf{Multi-Modal Attention.} At present, we employ self-attention to produce multi-modal hidden representations that fuse the information of question and video. In particular, we concatenate the question word-level features $Q = \{\mathbf{w}_{i}\}_{i=0}^{M-1}$, selected segment features $S_{\text{st}} = \{\mathbf{s}_{b} \; | \; b \in \mathcal{B} \}$, and selected patch features $H_{\text{st}} = \{\mathbf{h}_{\tau,j}|\tau \in \mathcal{T}\}_{j=0}^{N-1}$:
\begin{equation}
J = \left[\text{Linear}(S_{\text{st}}), \text{Linear}(H_{\text{st}}), \text{Linear}(\mathbf{q})\right],
\end{equation}
where $[;]$ denotes the concatenation operator. Thereafter, we iteratively conduct self-attention over the concatenated features for $N_{L}$ layers to accomplish multi-modal contextual representations:
\begin{equation}
J^{(i+1)} = \text{SelfAttention} \left(J^{(i)}\right), \; 0 \leq i \leq N_{L} - 1.
\label{eq:multimodal_self_attention}
\end{equation}

\noindent\textbf{Answer Prediction.} Afterwards, we pool the features of all multi-modal self-attention layers:
\begin{equation}
J_{\text{o}} = \text{pool}\left(J^{(0)}, J^{(1)}, ..., J^{(N_{L}-1)}\right).
\end{equation}
Then, we calculate the cosine similarity between $J_{o}$ and the feature of all candidate answers $X_{\text{ans}} = \{\mathbf{x}_{a} \; | \; a \in \mathcal{A}\}$ obtained by utilizing the pre-trained model. We choose the candidate answer of the highest similarity as the final prediction $\tilde{y}$:
\begin{equation}
\tilde{y} = \argmax_{y \in \mathcal{A}} \left(J_{\text{o}} \cdot \left(X_{\text{ans}}\right)^{\top}\right).
\end{equation}



\subsection{Cross-modal Compositional Congruence (C$^3$) Objective} 
Based on Eq. (\ref{eq:multimodal_self_attention}), we denote the output representations of visual patches as $J_{\text{v}}$ and those of question words as $J_{\text{w}}$. We calculate cross-modal attention:
\begin{gather}
G_{\text{vw}} = J_{\text{v}} \cdot \left(J_{\text{w}}\right)^{\top}, \;\; G_{\text{wv}} = J_{\text{w}} \cdot \left(J_{\text{v}}\right)^{\top},
\end{gather}
and intra-modal visual and textual attention as:
\begin{gather}
G_{\text{vv}} = J_{\text{v}} \left(J_{\text{v}}\right)^{\top}, \;\; G_{\text{ww}} =  J_{\text{w}} \left(J_{\text{w}}\right)^{\top}.
\end{gather}

Given the intra-modal and cross-modal attention, we perform the change of basis to compute the intra-modal visual attention in the language space:
\begin{gather}
R_{\text{vv}} = G_{\text{vw}} G_{\text{ww}}  \left(G_{\text{vw}}\right)^{\top}.
\end{gather}
As such, we define the loss objective to align the original $G_{\text{vv}}$ with the change-of-basis version $R_{\text{vv}}$:
\begin{equation}
\mathcal{L}_{\text{C}^{3}} = \text{m-KL}\left(\text{softmax}\left(R_{\text{vv}}\right), \text{softmax}\left(G_{\text{vv}}\right)\right), 
\end{equation}
where m-KL denotes the symmetric Kullback-Leilber Divergence (KL) between $R$ and $G$:
\begin{equation}
\text{m-KL}(R,G) = \text{KL}(R||G) + \text{KL}(G||R).
\end{equation}

\subsection{Overall Objective}
We jointly optimize the softmax cross-entropy loss $\mathcal{L}_{\text{CE}}$ between the log-likelihood of the model prediction $\tilde{y}$ and the groundtruth answer $y$, with our proposed cross-modal alignment objective $\mathcal{L}_{\text{C}^{3}}$:
\begin{equation}
\mathcal{L} = \mathcal{L}_{\text{CE}} + \gamma \cdot \mathcal{L}_{\text{C}^{3}},
\end{equation}
where $\gamma$ denotes the hyperparameter to control the effect of the cross-modal alignment objective.

\section{Ego-QA and MAD-QA Benchmarks}
To more rigorously evaluate long-form videoQA models, we construct two new datasets, \textit{\textit{i.e.}} Ego-QA and MAD-QA.

\subsection{Ego-QA} 
We inherit 3k hours of 8640 egocentric videos from the Ego4D dataset \citep{grauman2022ego4d}. Each video is associated with about 280 dense captions of consecutive moments. Based on these captions, we create our dataset in 2 stages, \textit{\textit{i.e.}} question-answer generation and data filtering. 

\noindent\textbf{Question-answer generation.} In this first stage, we concatenate a video’s dense captions following the time order to construct its language description. We utilize GPT-4 \citep{achiam2023gpt} to generate 20 questions per video. In our prompt, we encourage GPT-4 to avoid questions that are visually biased and can be answered by a short video moment. Then, we present the generated questions to GPT-4 to generate the correct answer along with 4 wrong answer choices. 

\noindent\textbf{Data filtering.} In the second stage, we filter out questions that include clue words, \textit{e.g.} ``\textit{passage}'', ``\textit{text}'', and ``\textit{description}''. Moreover, we also remove questions that GPT-4 can answer without looking at the concatenated narration or the question. Then, we adopt manual filtering by asking ten graduate students who are native English speakers to ensure the veracity and temporal certificate length for every question-answer sample. Particularly, annotators are instructed to verify that 1) questions are valid and the correct answer is indeed correct, 2) all distractor answers are incorrect, and 3) the video length to watch to determine the correct answer is at least 2 minutes. 

The filtering stage reduces the number of admissible questions by a factor of $4\times$ to $5\times$. We accomplish 18.8K questions for 992 videos, which we split into 80\% train, 10\% val, and 10\% test. 

\subsection{MAD-QA} 
We follow the same process for Ego-QA to obtain MAD-QA by utilizing 1.2K hours of 650 videos from the MAD dataset \citep{soldan2022mad}. Since video lengths and the number of dense captions are larger than Ego-QA, we ask GPT-4 to generate 60 instead of 20 questions per video. Because GPT-4 might store external knowledge about the movie, we replace the name of characters in the caption with $person\_1$, $person\_2$, etc. Afterwards, we obtain 15.7K questions for 650 videos, and we split them into 80\%  train, 10\% val, and 10\% test.

The average video lengths in Ego-QA and MAD-QA are 17.5 minutes and 1.9 hours, respectively. Moreover, the average necessary video lengths humans need to watch to determine the answer for the two datasets are respectively 1204.4 and 396.07 seconds, longer than the average 100-second length of the recent very long-form videoQA dataset EgoSchema \citep{mangalam2023egoschema}. We visualize the statistics of our datasets in Appendix \ref{app:dataset_statistics}, and language prompts to generate question-answer samples in Appendix \ref{app:question_answer_prompts}, along with the instructions for human annotators in Appendix \ref{app:manual_annotation}.

\section{Experiments}
\begin{table*}[t]
\centering
\resizebox{\linewidth}{!}{
\begin{tabular}{l|c|c|c|c|c|c|c|c|c}
\hline
\multicolumn{1}{c|}{\textbf{Question Types}} & \textbf{Object-relation} & \textbf{Relation-action} & \textbf{Object-action} & \textbf{Superlative} & \textbf{Sequencing} & \textbf{Exists} & \textbf{Duration comparison} & \textbf{Activity recognition} & \textbf{All} \\ \hline
AIO                                         & 48.34                    & 48.99                    & 49.66                  & 37.53                & 49.61               & 50.81           & 45.36                        & 18.97                         & 48.59        \\
ATP                                         & 50.15                    & 49.76                    & 46.25                  & 39.78                & 48.25               & 51.79           & 49.59                        & 18.96                         & 49.79        \\
MIST-AIO                                    & 51.43                    & 54.67                    & 55.37                  & 41.34                & 53.14               & 53.49           & 47.48                        & 20.18                         & 50.96        \\
MIST-CLIP                                   & 51.68                    & 67.18                    & 68.99                  & 42.05                & 67.24               & 60.33           & 54.62                        & 19.69                         & 54.39        \\ \hline
GSMT-AIO                            &       53.67                   &           56.10               &                 56.61       &          43.44            &          53.84           &         56.26        &                            49.83   &                21.73               &       52.61       \\
GSMT-CLIP                &    \textbf{53.94}       &           \textbf{69.84}               &     \textbf{72.53}                        &            \textbf{44.19}           &           \textbf{69.12}             &      \textbf{61.45}              &          \textbf{57.32}                          &               \textbf{21.20}                &     \textbf{56.16}        \\ \hline
\end{tabular}}
\caption{Results of videoQA on AGQA-v2.}
\label{tab:exp_agqav2}
\end{table*}

\begin{table}[t]
\centering
\resizebox{\linewidth}{!}{
\begin{tabular}{l|ccccc|c}
\hline
\multicolumn{1}{c|}{\textbf{Method}} & \textbf{Attribute} & \textbf{State} & \textbf{Event} & \textbf{Order} & \textbf{Number} & \textbf{All} \\ \hline
STAGE                               & 39.49              & 49.93          & 34.52          & 55.32          & 38.54           & 41.97        \\
AIO                                 & 41.78              & 52.98          & 37.57          & 55.16          & 38.50           & 44.86        \\
ATP                                 & 42.87              & 53.49          & 38.35          & 55.25          & 38.65           & 45.43        \\
MIST-AIO                            & 43.63              & 55.17          & 40.99          & 55.44          & 39.54           & 47.19        \\
MIST-CLIP                           & 44.05              & 58.13          & 42.54          & 56.83          & 40.32           & 48.97        \\ \hline
GSMT-AIO                          &      48.76              &       57.99         &             44.96    &           57.39      &    43.18            &       50.81       \\
GSMT-CLIP                          &        \textbf{49.23}            &     \textbf{61.10}           & \textbf{46.66}               &     \textbf{58.83}           &       \textbf{44.03}          &     \textbf{52.73}       \\ \hline 
\end{tabular}}
\caption{Results of videoQA on Env-QA.}
\label{tab:exp_envqa}
\end{table}

\begin{table}[t]
\centering
\resizebox{\linewidth}{!}{
\begin{tabular}{l|cccc|c}
\hline
\multicolumn{1}{c|}{\textbf{Method}} & \textbf{Interaction} & \textbf{Sequence} & \textbf{Prediction} & \textbf{Feasibility} & \textbf{Mean} \\ \hline
CLIP                                & 39.80                & 40.50             & 35.50               & 36.00                & 38.00         \\
RESERVE-B                           & 44.80                & 42.40             & 38.80               & 36.20                & 40.50         \\
Flamingo-9B                         & -                    & -                 & -                   & -                    & 43.40         \\
AIO                                 & 47.53                & 50.81             & 47.75               & 44.08                & 47.54         \\
ATP                                 & 50.63                & 52.87             & 49.36               & 40.61                & 48.37         \\
CoVGT                           & -               & -             & -               & -                & 46.23         \\
MIST-AIO                            & 53.00                & 52.37             & 49.52               & 43.87                & 49.69         \\
MIST-CLIP                           & 55.59                & 54.23             & 54.24               & 44.48                & 51.13         \\ \hline
GSMT-AIO                    &          56.59            &          55.55         &         52.23            &             46.04        &       51.36        \\
GSMT-CLIP                   &             \textbf{59.36}         &          \textbf{57.52}         &            \textbf{57.21}         &          \textbf{46.68}            &     \textbf{52.85}         \\ \hline
\end{tabular}}
\caption{Results of videoQA on STAR.}
\label{tab:exp_star}
\end{table}

\begin{table}[t]
\centering
\resizebox{0.9\linewidth}{!}{
\begin{tabular}{l|ccc|c}
\hline
\multicolumn{1}{c|}{\textbf{Method}} & \textbf{Causal} & \textbf{Temporal} & \textbf{Descriptive} & \textbf{All} \\ \hline
HQGA                                & 48.48           & 51.24             & 61.65                & 51.42        \\
CLIP                                & 46.30           & 39.00             & 53.10                & 43.70        \\
VQA-T                               & 49.60           & 51.49             & 63.19                & 52.32        \\
AIO                                 & 48.04           & 48.63             & 63.24                & 50.60        \\
ATP                                 & 53.10           & 50.20             & 66.80                & 54.30        \\
CoVGT                               & 59.69           & 58.00             & 69.88                & 60.73        \\
MIST-AIO                            & 51.54           & 51.63             & 64.16                & 53.54        \\
MIST-CLIP                           & 54.62           & 56.64             & 66.92                & 57.18        \\ \hline
GSMT-AIO                    &           59.72      &        59.04           &         69.91             &         60.76     \\
GSMT-CLIP                   &         \textbf{60.87}        &         \textbf{61.16}          &           \textbf{70.26}           &         \textbf{62.49}     \\ \hline
\end{tabular}}
\caption{Results of videoQA on NExT-QA.}
\label{tab:exp_nextqa}
\end{table}

\begin{table}[t]
\centering
\resizebox{0.45\linewidth}{!}{
\begin{tabular}{l|c}
\hline
\multicolumn{1}{c|}{\textbf{Method}} & \textbf{Accuracy} \\ \hline
EgoVLP                              & 34.86             \\
VideoReCap                 & 50.23             \\
MIST-AIO                            & 56.27             \\
MIST-CLIP                           & 56.42             \\ \hline
GSMT-AIO                    & 58.28             \\
GSMT-CLIP                   & \textbf{58.55}            \\ \hline
\end{tabular}}
\caption{Results of videoQA on EgoSchema.}
\label{tab:exp_egoschema}
\end{table}

\begin{table}[t]
\centering
\resizebox{0.6\linewidth}{!}{
\begin{tabular}{l|c|c}
\hline
\multicolumn{1}{c|}{\textbf{Method}} & \textbf{Ego-QA} & \textbf{MAD-QA} \\ \hline
AIO                                 & 24.19           & 16.14           \\
CoVGT                               & 26.72           & 15.71           \\
MIST-AIO                            & 27.71           &  14.19           \\
MIST-CLIP                           & 29.73           &   17.15         \\  \hline
GSMT-AIO                    &       28.72           &      15.69           \\
GSMT-CLIP                   &     \textbf{32.40}            &   \textbf{19.11}
\\ \hline    
Human                           &    80.29      &   73.21         \\ \hline
\end{tabular}}
\caption{Results on constructed Ego-QA and MAD-QA.}
\label{tab:exp_egoqa_madqa}
\end{table}

\begin{table}[t]
\centering
\resizebox{\linewidth}{!}{
\begin{tabular}{l|c|c|c|c}
\hline
\multicolumn{1}{c|}{\textbf{Method}} & \textbf{NExT-QA} & \textbf{STAR} & \textbf{Ego-QA} &  \textbf{MAD-QA} \\ \hline
SSL                            &  61.91     & 52.40            & 31.30                & 18.95           \\
Non-diag SSL                     &  60.44           & 51.63            & 30.51                  & 18.86           \\
Attention                 &   59.74             & 51.07            & 30.06                  & 18.41  \\ 
Convolution                 &   59.34             & 50.85            & 30.17                  & 18.33  \\ \hline
Gated SSL                 &   \textbf{62.49}            & \textbf{52.85}           & \textbf{32.40}                 & \textbf{19.11} \\ \hline
\end{tabular}}
\caption{Ablation results of gated SSL.}
\label{tab:ablation_gss}
\end{table}

\begin{table}[t]
\centering
\resizebox{0.95\linewidth}{!}{
\begin{tabular}{l|c|c|c|c}
\hline
\multicolumn{1}{c|}{\textbf{Method}} & \textbf{NExt-QA} & \textbf{STAR} & \textbf{Ego-QA} & \textbf{MAD-QA} \\ \hline
GSMT           &      60.78      & 49.67            & 29.73         &      18.00    \\
GSMT w/ OT        &   60.89      & 50.30            & 30.14         & 18.22        \\  
GSMT w/ POT       &     60.93    & 50.59            & 30.55         & 18.26        \\  \hline 
GSMT w/ C$^3$      &    \textbf{62.49}      & \textbf{52.85}           & \textbf{32.40}       & \textbf{19.11}        \\  
\hline
\end{tabular}}
\caption{Ablation results of cross-modal alignment.}
\label{tab:ablation_cross_modal_alignment}
\end{table}

\begin{table}[t]
\centering
\resizebox{\linewidth}{!}{
\begin{tabular}{l|c|c|c|c|c|c}
\hline
\multicolumn{1}{c|}{\textbf{Dataset / $d_{\text{gating}}$}} & \textbf{32} & \textbf{64} & \textbf{128} & \textbf{256} & \textbf{512} & \textbf{No gating} \\ \hline
NExT-QA                                                    & 60.74       & 61.09       & \textbf{62.49}        & 61.90        & 61.33        & 60.61              \\
STAR                                                       & 51.82       & 52.63       & \textbf{52.85}        & 52.70        & 51.86        & 51.27              \\
Ego-QA                                                     & 29.96       & 31.37       & \textbf{32.40}        & 31.55        & 31.25        & 29.88   \\ \hline
\end{tabular}}
\caption{Effect of various $d_{\text{gating}}$ dimensions and no gating on NExT-QA, STAR, and Ego-QA datasets.}
\label{tab:ablation_study_gating_unit}
\end{table}

\begin{table}[h!]
\centering
\resizebox{\linewidth}{!}{
\begin{tabular}{l|c|c|c|c}
\hline
\multicolumn{1}{c|}{\textbf{Method}} & \textbf{NExT-QA} & \textbf{STAR} & \textbf{Ego-QA} & \textbf{MAD-QA} \\ \hline
Multi-modal SSL                     & 60.69            & 50.87         & 29.74           & 18.61           \\
GSMT                        & \textbf{62.49}            & \textbf{52.85}        & \textbf{32.40}           & \textbf{19.11}           \\ \hline
\end{tabular}}
\caption{Effect of the position of SSL.}
\label{tab:ablation_study_position_ssl}
\end{table}

\subsection{Standard Benchmarks}
In addition to our constructed Ego-QA and MAD-QA datasets, we follow previous works \citep{gao2023mist, xiao2023contrastive, wang2023all} to evaluate GSMT on four additional publicly available datasets for long-form videoQA: AGQA \citep{grunde2021agqa}, NExT-QA \citep{xiao2021next}, STAR \citep{wu2021star}, Env-QA \citep{gao2021env}, and EgoSchema \citep{mangalam2023egoschema}.

\noindent\textbf{AGQA} \citep{grunde2021agqa} is a videoQA dataset for compositional spatio-temporal reasoning. As recommended by the dataset creator, we employ its v2 version, which possesses more balanced distributions. AGQA consists of 2.27M QA pairs for 9.7K videos.

\noindent\textbf{NExT-QA} \citep{xiao2021next} focuses on causal and temporal reasoning. The dataset comprises 5,440 videos associated with 52K questions.

\noindent\textbf{STAR} \citep{wu2021star} concentrates on situated reasoning questions. The dataset provides 60K questions related to 22K videos clips.

\noindent\textbf{Env-QA} \citep{gao2021env} is curated for dynamic environment understanding. Env-QA contains 23K egocentric videos collected on virtual environment AI2THOR \citep{kolve2017ai2}, which are used to generate 85K questions.

\noindent\textbf{EgoSchema} \citep{mangalam2023egoschema} consists of egocentric videos of 3-minute length. Questions in EgoSchema require humans on average 100 seconds watching the video to answer. 

\begin{figure}[t]
    \centering
    \includegraphics[width=\linewidth]{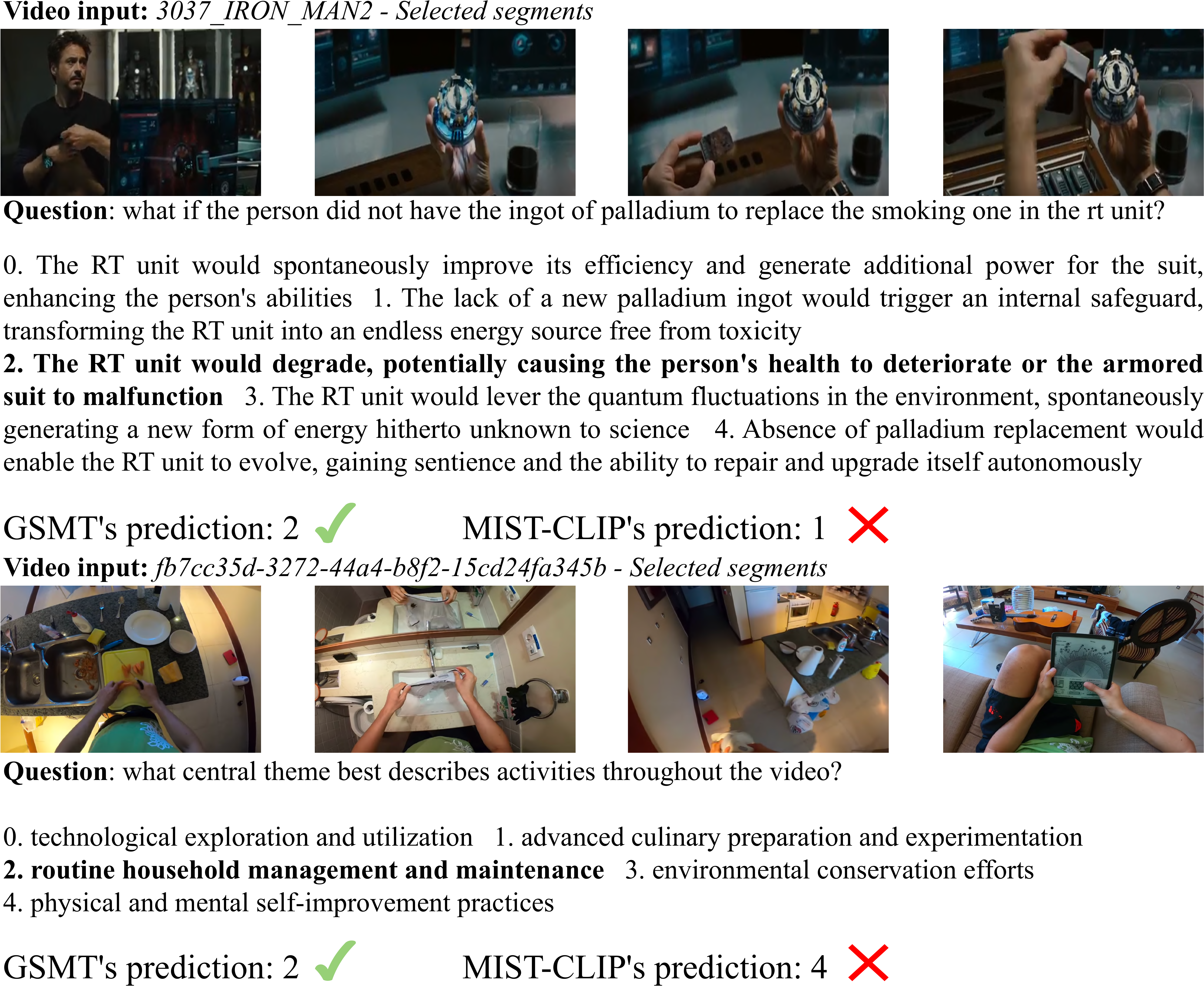}
    \caption{Qualitative results on the constructed MAD-QA and Ego-QA datasets.}
    \label{fig:example_qualitative_analysis}
\end{figure}

\begin{figure}[t]
    \centering
    \includegraphics[width=0.75\linewidth]{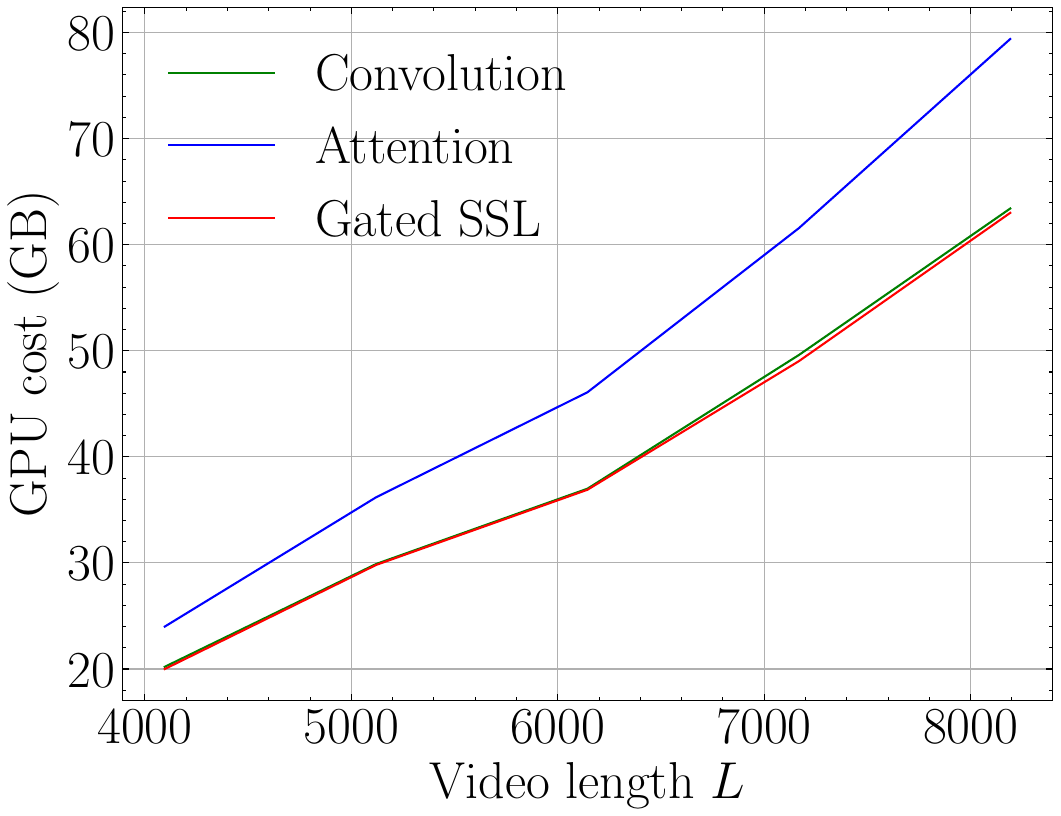}
    \caption{GPU memory cost with respect to visual sequence length $L$ of attention, convolution, and gated SSL mechanism.}
    \label{fig:ablation_study_gpu_cost}
\end{figure}

\subsection{Implementation Details}
Our framework can be implemented on most multi-modal Transformers. To fairly compare with previous works \citep{wang2023all, gao2023mist}, we evaluate upon two popular types of pre-trained models, \textit{\textit{i.e.}} CLIP (ViT/B-32) \citep{radford2021learning} and All-in-One-Base \citep{wang2023all}. We divide a video frame into $4\times4$ patches and send them to video embedder. For our gated SSL, we use $d_S = d = d_{h} = 512, d_{\text{gating}} = 128$. For selection modules, we use top-$k$ = 4, top-$j$ = 12. For the multimodal attention module, we use $N_L = 2$. Based on validation, we employ max-pooling for all pooling operations, ReLU activation for $\phi$, and $N_L = 2$. We apply $\mathcal{L}_{C^{3}}$ upon $J^{(2)}$, and observe no difference between applying on $J^{(1)}$ and $J^{(2)}$.  For fair comparison on AGQA, NExT-QA, STAR, Env-QA, and EgoSchema datasets, we sample 32 frames per video, and split them into $K = 8$ segments. Since video lengths are longer in our EgoQA and MAD-QA datasets, we sample 128 and 8192 frames per video, respectively, and split into $K = 8$ segments. For language modality, we embed the question with the same pre-trained model as the video embedder, and embed the answer with the pre-trained BERT-base model. We apply $\lambda = 0.005$ to balance the scale of $\mathcal{L}_{C^{3}}$ and $\mathcal{L}_{\text{CE}}$.


\subsection{Baselines}
We compare our proposed framework against the following baseline:
\begin{itemize}
    \item \textbf{HQGA} \citep{xiao2022hqga}: models video as a conditional graph hierarchy which aggregates visual facts in a level-wise manner with the guidance of textual modality,
    \item \textbf{CoVGT} \citep{xiao2023contrastive}: a model which exploits a graph transformer to encode video by capturing visual objects and their relations for spatio-temporal reasoning, along with contrastive learning to align its graph transformer with the textual encoder,
    \item \textbf{ATP} \citep{buch2022revisiting}: a model which learns to select one salient video frame to perform the videoQA task,
    \item \textbf{AIO} \citep{wang2023all}: an end-to-end videoQA model that releases the need of unimodal encoders and leverages non-parametric token rolling operation,
    \item \textbf{STAGE} \citep{lei2020tvqa+}: a model that jointly carries out videoQA with moment localization and object grounding,
    \item \textbf{VQA-T} \citep{yang2021just}: a cross-modal Transformer which is pre-trained with supervised contrastive learning on a large-scale videoQA dataset,
    \item \textbf{RESERVE-B} \citep{zellers2022merlot}: jointly represents video frames, texts, and audio, and learns to predict the masked textual and audio tokens given video frames,
    \item \textbf{Flamingo-9B} \citep{alayrac2022flamingo}: a model which is able to process multi-modal prompt and cast videoQA as text prediction,
    \item \textbf{EgoVLP} \citep{lin2022egocentric}: pre-trains multi-modal Transformer on egocentric videos for egocentric videoQA.
    \item \textbf{VideoReCap} \citep{islam2024video}: a recursive model that synergizes different video hierarchies to process hour-long videos,
    \item \textbf{MIST} \citep{gao2023mist}: a multi-modal Transformer which decomposes video into frames, patches, and segments to efficiently process with its self-attention mechanism.
\end{itemize}
\subsection{Quantitative Results}
We show our experiments on AGQA-v2, Env-QA, STAR, NExT-QA, and EgoSchema in Table \ref{tab:exp_agqav2}, \ref{tab:exp_envqa}, \ref{tab:exp_star}, \ref{tab:exp_nextqa}, and \ref{tab:exp_egoschema}, respectively, while revealing results on our constructed Ego-QA and MAD-QA in Table \ref{tab:exp_egoqa_madqa}. We can observe that our method achieves superior performance over the latest methods on all datasets. In terms of the overall accuracy, we outperform the second-best method on AGQA-v2, Env-QA, STAR, and EgoSchema, \textit{\textit{i.e.}} MIST-CLIP, by 1.77\%, 3.76\%, 1.72\%, and 2.13\%, respectively. Equivalently, we outperform CoVGT, which is the second-best method on NExT-QA, by 1.20\%.

Inspecting more closely, we note that our framework obtains more significant performance increase on questions that require the capacity of reasoning among visual concepts, \textit{\textit{i.e.}} improving 2.66\% and 3.54\% respectively for \textit{relation-action} and \textit{object-action} on AGQA-v2, 6.25\% and 4.52\% respectively for causal and temporal on NExT-QA, than those that require the ability to extract information within one frame, \textit{\textit{i.e.}} improving 2.26\% for \textit{object-relation} on AGQA-v2 and 3.34\% for \textit{descriptive} on NExT-QA. These results demonstrate our global semantics signal can address the challenging long-range temporal reasoning problems of long-form videoQA.

Remarkably, existing methods demonstrate significantly low performance on our curated datasets. For example, MIST-CLIP only achieves 29.73\% on Ego-QA, and 17.15\% accuracy on MAD-QA, which is less than random chance. In contrast, humans obtain 80.29\% and 73.21\% accuracy on Ego-QA and MAD-QA, respectively. These results suggest that previous methods might not encompass sufficient information in their selected segments and visual regions. Conversely, with the integrated global information, our framework can enhance videoQA performance on these challenging datasets. However, the accuracy remains substantially below human performance. Future research should focus more on genuine long-form videoQA, where videos can extend to several hours.

\subsection{Ablation Study}

\noindent\textbf{Gated SSL implementation.} We explore the effect of our gated SSL in Table \ref{tab:ablation_gss} on NExT-QA, STAR, Ego-QA, and MAD-QA datasets. As can be observed, removing the gating unit, \textit{i.e.} the SSL approach, results in performance drops, since redundant and noisy information might be passed to the visual representations. Additionally, not initializing state space parameters as diagonal matrices, \textit{i.e.} non-diag SSL, does not remarkably impact the performance. However, the time and memory complexity would become $O(L^2)$, which is significantly more costly than our initialization approach.

\noindent\textbf{Choices for Global Semantics.} We compare gated SSL with other choices to extract global semantics among visual elements, \textit{i.e.} self-attention and convolution, in terms of videoQA performance in Table \ref{tab:ablation_gss} and GPU memory cost in Figure \ref{fig:ablation_study_gpu_cost}. As can be observed, our gated SSL not only brings less computational cost than self-attention but also higher accuracy, validating the effectiveness of its global information signal. Moreover, whereas convolution pays equivalent computational cost to our gated SSL, its local pattern does not provide productive contextual information among visual elements, resulting in lower accuracy than gated SSL. 

\noindent\textbf{Effect of Gating Unit.} We ablate the gating unit and vary the gating dimension $d_{\text{gating}}$ in our gated SSL. As shown in Table \ref{tab:ablation_study_gating_unit}, increasing the gating dimension leads to higher videoQA accuracy, as model has more controllability towards global information into visual representations. 
However, when the gating dimension becomes larger, the performance saturates and deteriorates. We posit that the model might become more constrained to allow the encoding of global semantics, degenerating to the architecture with limited global semantics. Apparently, removing gating unit results in performance drops, because irrelevant global information for the question could flow into visual hidden states without model controllability.

\noindent\textbf{Effect of C$^3$ objective.} We compare our $C^{3}$ alignment objective with alternative approaches, \textit{i.e.} optimal transport (OT) \citep{pramanick2022multimodal} and its partial variant (POT) \citep{chapel2020partial}. As shown in Table \ref{tab:ablation_cross_modal_alignment}, both OT and POT can polish videoQA performance, while C$^3$ yields the highest performance. This shows that compositional consistency is important for long-form videoQA, since the model needs to grasp the relations among entities, specifically those specified by the question.

\noindent\textbf{Position of State Space Layer.} We replace the penultimate multi-modal attention with SSL. As shown in Table \ref{tab:ablation_study_position_ssl}, such design choice deteriorates the videoQA performance. The reason might be that since the frames and regions have already been selected, limited global information can be extracted from the video. Moreover, SSL does not explicitly calculate dependency between tokens, thus producing little refined representations to compute the final answer, which has been observed by previous work \citep{zuo2022efficient}. This substanstiates our decision to adopt SSL to integrate global semantics of video in an earlier stage. 

\subsection{Qualitative Results}

We visualize videoQA cases in Figure \ref{fig:longform_videoqa_example}. As can be observed, our model can choose the correct answer for questions that require information over the video with a limited number of video segments. We posit that due to our integrated global semantics, visual representations not only encode information of the selected segments but also the video context, thus furnish the model with sufficient cues to ascertain the correct answer. In contrast, constrained to the selected segments, previous state-of-the-art, \textit{i.e.} MIST-CLIP \citep{gao2023mist}, struggles in such questions and produces the incorrect output.
\section{Conclusion}
We introduce a Gated State space Multi-modal Transformer (GSMT) with a state space layer (SSL) to integrate global semantics of video into visual representations to tackle long-form videoQA. We further incorporate a gating unit to provide more controllability over the integrated global semantics and a cross-modal compositional congruence (C$^3$) objective to encourage the semantics aligned with the question. To comprehensively evaluate long-form videoQA, we curate two long-form videoQA datasets with excessively long video lengths and long-natured questions. Extensive experiments on these and standard datasets validate the superiority of our framework.
\section{Limitations}
Our proposed framework has achieved promising improvement by integrating productive global context of long videos for long-form videoQA, but we consider the following limitations as future work:
\begin{itemize}[leftmargin=*]
    \item \textbf{Improve the generalizability of videoQA systems:} Long videos exhibit diverse content, which make it unlikely that the model will encounter inputs of the same distribution during inference. Since training different models for different settings is computationally costly, it is desirable to construct a long-form videoQA model that can generalize to myriad content.

    \item \textbf{Extend our datasets to multicultural settings:} Many long videos from different nations with distinct backgrounds are prevalent in the Internet. Therefore, to increase the usefulness of long-form videoQA model, there is a need to construct benchmarks for different cultures and societies to facilitate the development of future multi-cultural long-form videoQA systems.
\end{itemize}

\section*{Acknowledgement} 
This research/project is supported by the National Research Foundation, Singapore under its AI Singapore Programme (AISG Award No: AISG3-PhD-2023-08-051T). Thong Nguyen is supported by a Google Ph.D. Fellowship in Natural Language Processing.

\bibliography{anthology, custom}

\appendix
\onecolumn
\section{Prompt for generating Ego-QA and MAD-QA datasets}
\label{app:question_answer_prompts}
\subsection{Question prompt}
\begin{lstlisting}
I want you to act as a teacher in the class called ``Long-term video understanding''. I will provide video narrations describing events in the time order of the video and you will generate highly difficult and diverse questions for your students about the high-level details in the video. You want to test students' following abilities: 

Ability 1: Students' ability to summarize and compare long parts of the video
Ability 2: Students' ability to compress information from the video rather than just listing the actions that happened in the video.
Ability 3: Students' ability to identify the most important parts of the video. 

Your questions should not mention any particular timestamps or narrations. Remember to make sure the correct answers to your questions do not list information from the narrations but compress them in a concise conclusion. 

Examples of good and difficult questions:
"What if A happened instead of ...?" 
"Why did A do action ...?"
"What did A do after/before ...?"

AVOID the following types of questions:
"When ...?"
"How many ...?"
"How much ...?"
When announcing the question please label each question as "Question 1,2,3: [full question]" 

Video narrations: [video narrations]
\end{lstlisting}

\subsection{Answer prompt}
\begin{lstlisting}
I want you to act as a teacher in the class called ``Long-term video understanding.'' I will provide video action narrations and highly difficult and diverse questions for your students about the high-level details in the video. I want you to test students' following abilities:

Ability 1: Students' ability to summarize and compare long parts of the video
Ability 2: Students' ability to compress information from the video rather than just
listing the actions that happened in the video.
Ability 3: Students' ability to identify the most important parts of the video. 

I want you to create a difficult multiple-choice exam that tests above student abilities based on the questions I just provided. Each question should have five similar open-ended but short answers, but only one should be correct. Make it very difficult for students to find the correct answer among all the wrong answers. All answers should be closely related to what happens in the video. Make wrong answers significantly longer than correct answers. Ensure all of the correct answers compress information from narrations them into a concise conclusion. Your answers should not mention any particular timestamps or narrations. 

Do not use letters for the answer choices
Print each correct answer exactly as "Correct answer: [full answer]"
Please print each wrong answer on a new line and print each wrong answer as "Wrong answer 1,2,3,4: [full answer]" 

Video narrations: [video narrations]
Questions: [question list]
\end{lstlisting}

\section{Manual annotation}
\label{app:manual_annotation}
We utilize ten professional English speakers to ensure the quality of our datasets. Our manual annotation consists of two procedures: question filtering and human accuracy testing.

\subsection{Question filtering}
Human annotators are responsible for ensuring that the question-answer samples in two datasets are of the highest quality possible. We provide the instruction that we show to the annotators:

\begin{lstlisting}
The annotation we need is to say that the Question-correct answer-wrong answer set (the whole set) is good if all these three conditions pass:

(Condition A) Question is Answerable: The question can be answered from the video and requires more than just a second of video to answer (so, if the answer is not present in the video or, if the answer can be formed with just a few frames (less than say, a second) then it fails this condition). 

(Condition B) The Marked Correct Answer is correct: The ""correct answer"" is more the correct answer to the question

(Condition C) The Marked Wrong Answers are wrong: All 4 ""wrong answers"" are less correct than the ""correct answer"" (So for example, if a wrong answer is not completely false, but simply does not contain all the information that the "" correct answer"" does, then it is still a fine ""wrong answer"") IF even one of the marked answer is correct, the set should be labeled as bad.

(Condition D) The question is actually long-term: This is a very very important condition. We want the certificate for the question to be at least 30 seconds minimum. If the certificate is non-contiguous (I.e. 5 seconds at one place, 20 seconds at another, and 15 more seconds at a third place) the sum of lengths of all the sub-certificates together should be more than 30 seconds. Another example is, if a question can be answered simply from a few frames of the video, the certificate is small (and less than 30 seconds) and hence would fail this condition. Additional details on how to handle certificate edge cases are provided in the annotator training through examples.

(Condition E) Avoid Boring Questions: Questions that ask about the frequency of
something ("How many times..") fail this condition. 

If any of these five conditions fail we want the whole set (Question / Correct Answer / Wrong Answer) marked bad.

Optional:
Since GOOD sets are so rare, in cases where it seems that a set is good but a small
part of the above five conditions is not being met or, if one/two words were
different this can be a good set, please label as MAYBE and we will fix it in
the second round. 

Extended notes:
1. In our experience, the wrong answers are made such that they differ from the correct answer in small but crucially meaningful ways. There are many cases where a careless reading of the wrong answer might make it seem that it is correct but upon careful inspection, it will become clear that something about the wrong answer indeed makes it wrong. While this is common, there are indeed cases where the wrong answer is also just as correct as the correct answer. In such cases, the wrong answer fails condition C, and the set is bad.

2.Roughly speaking, we expect about 20-25% of the questions that we have provided to be found as good. However, this is not necessary and the percentage can be smaller or larger depending on each three-minute clip.

3. Edge Cases:

1. If the asked question has multiple answers and at least one of them aligns with the correct answer while none of them align with any of the other wrong answers, then provided that the top 5 conditions are met, we can mark the set as good.

2. If two questions are very similar (even within different clips) and both are GOOD, only choose one as GOOD and reject the other one with a comment mentioning this. We do not expect this to happen more than 1 or 2 times in a 100.

3. There might be more such edge cases, please feel free to contact me in such cases
and we can explain. 
\end{lstlisting}

\subsection{Human accuracy testing}
To benchmark human, we recruit another team of ten human annotators to answer the questions in the curated datasets. The answers are randomly shuffled and presented in the form of a test. 
\newpage

\section{Examples of Ego-QA dataset}
\label{app:example_ego_qa}
In this appendix, we provide more examples of our constructed Ego-QA dataset. Our videos are taken from the Ego4D dataset \citep{grauman2022ego4d}.
\begin{figure*}[h!]
    \centering
    \includegraphics[width=0.9\linewidth]{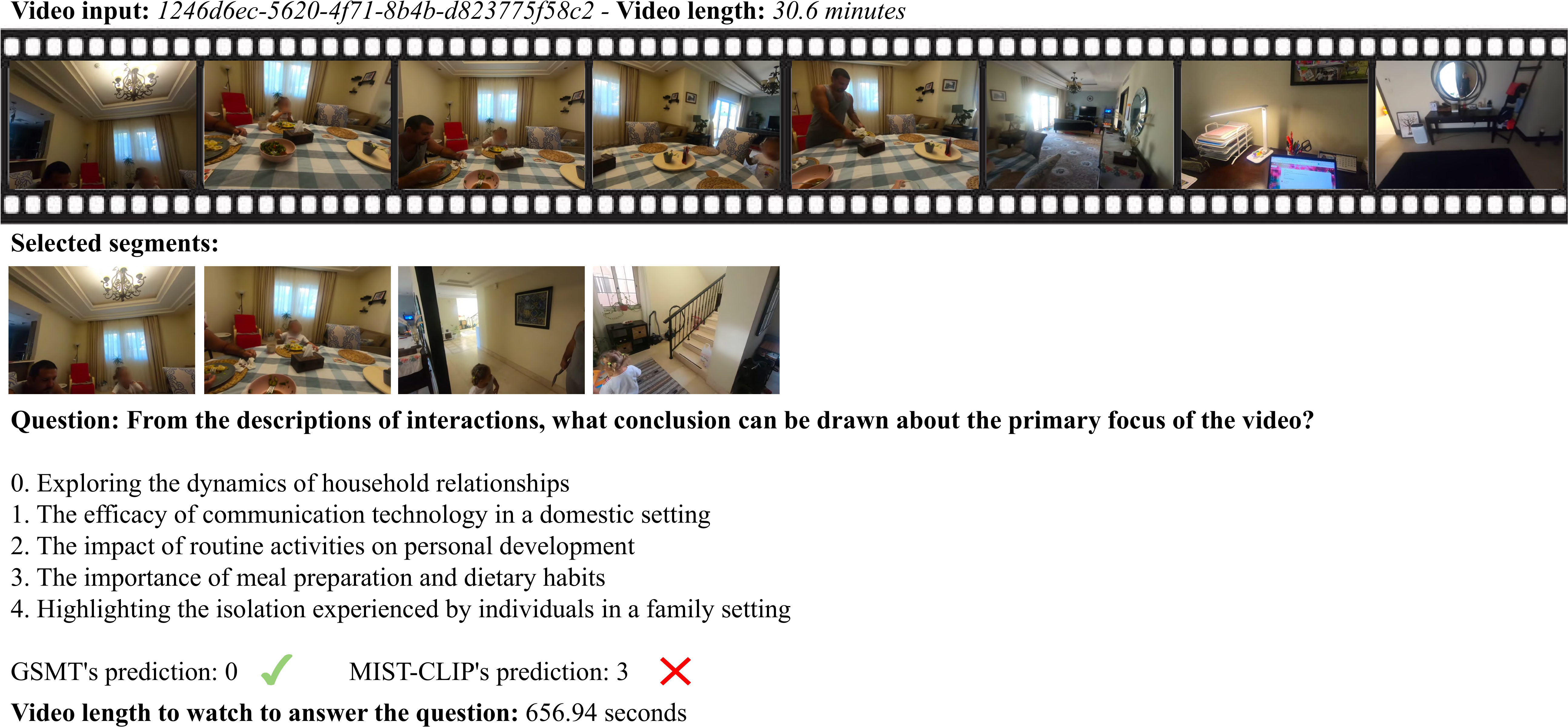}
\end{figure*}
\begin{figure*}[h!]
    \centering
    \includegraphics[width=0.9\linewidth]{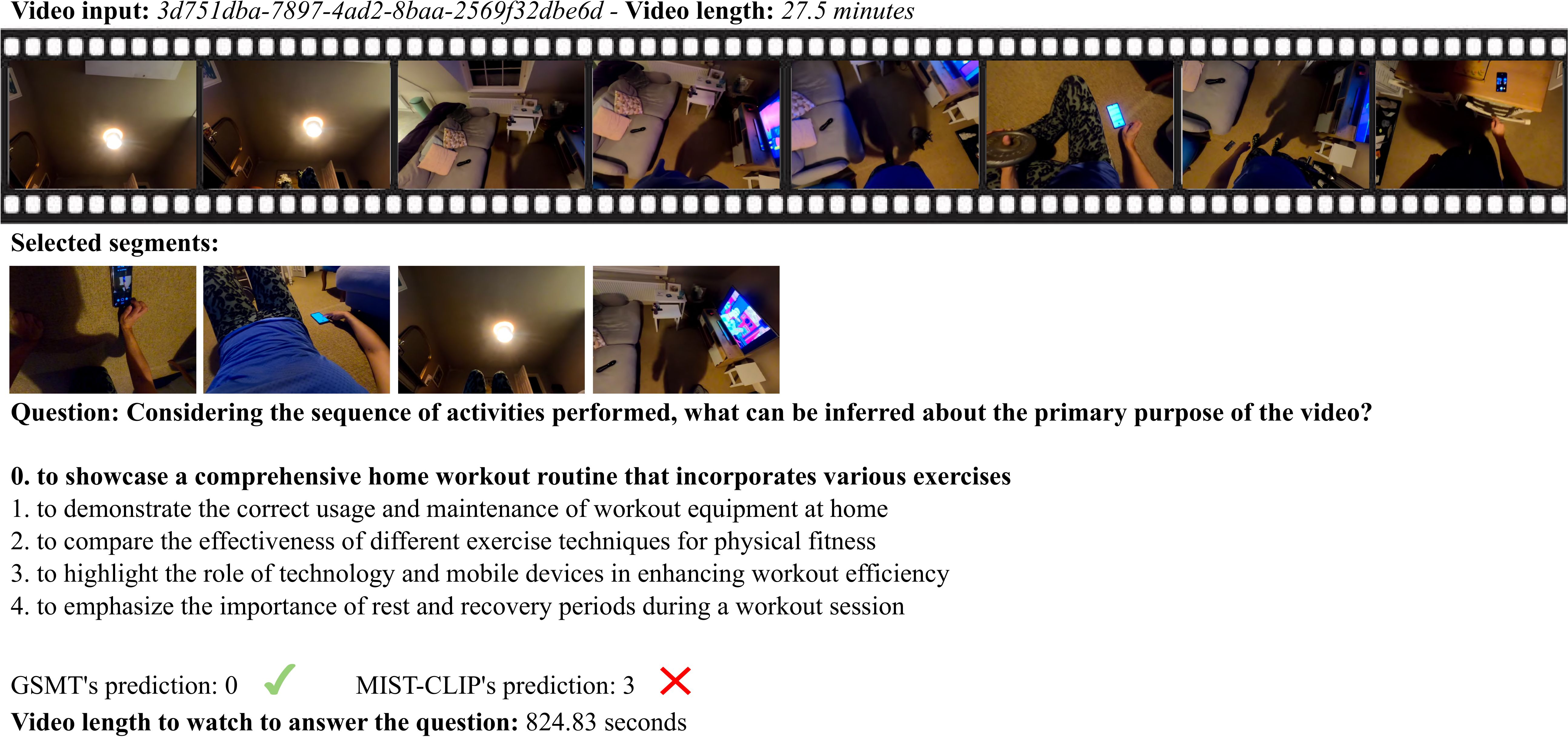}
\end{figure*}
\begin{figure*}[h!]
    \centering
    \includegraphics[width=0.9\linewidth]{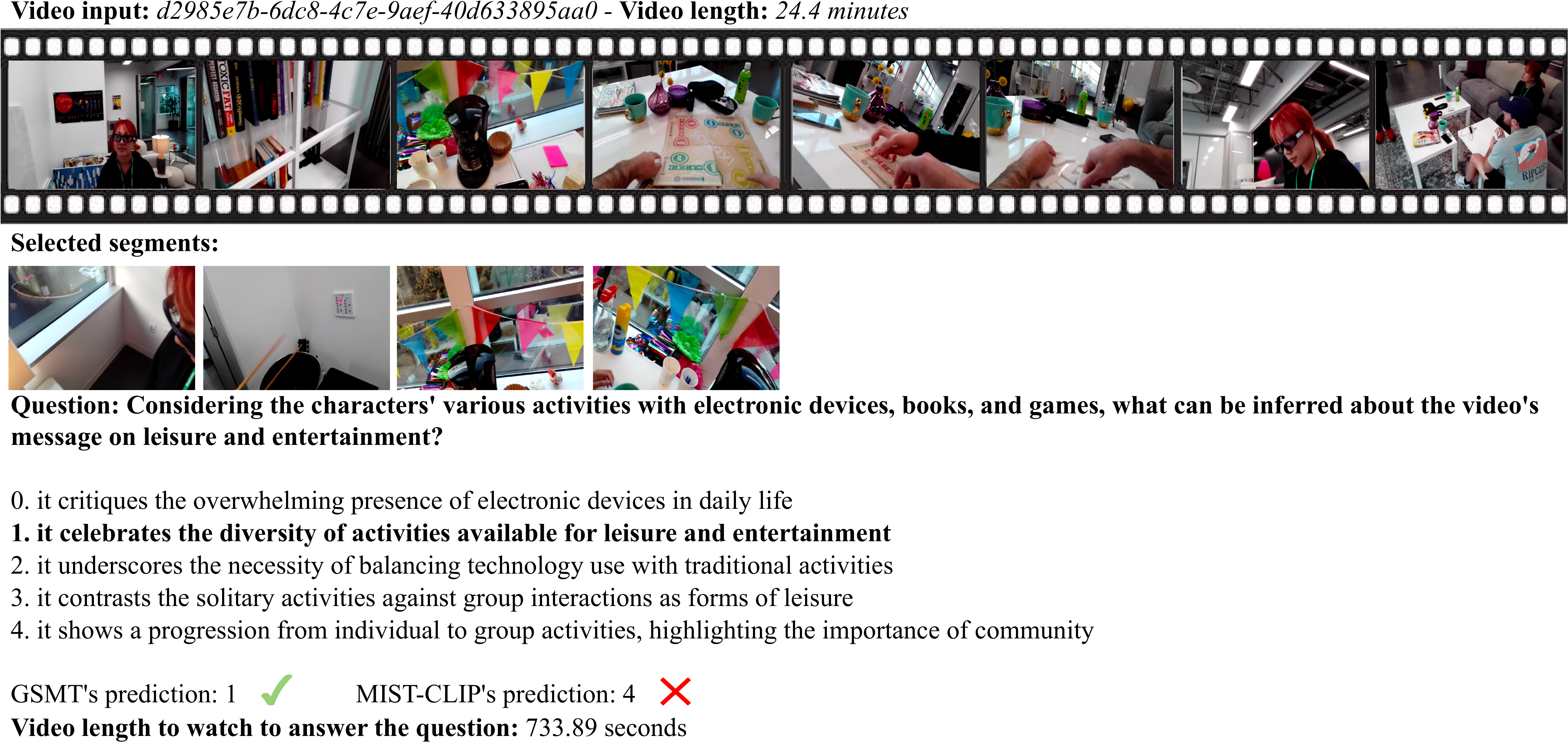}
\end{figure*}

\section{Examples of MAD-QA dataset}
\label{app:example_mad_qa}
In this appendix, we provide more examples of our constructed MAD-QA dataset. Our videos are taken from the MAD dataset \citep{soldan2022mad}.
\begin{figure*}[h!]
    \centering
    \includegraphics[width=0.9\linewidth]{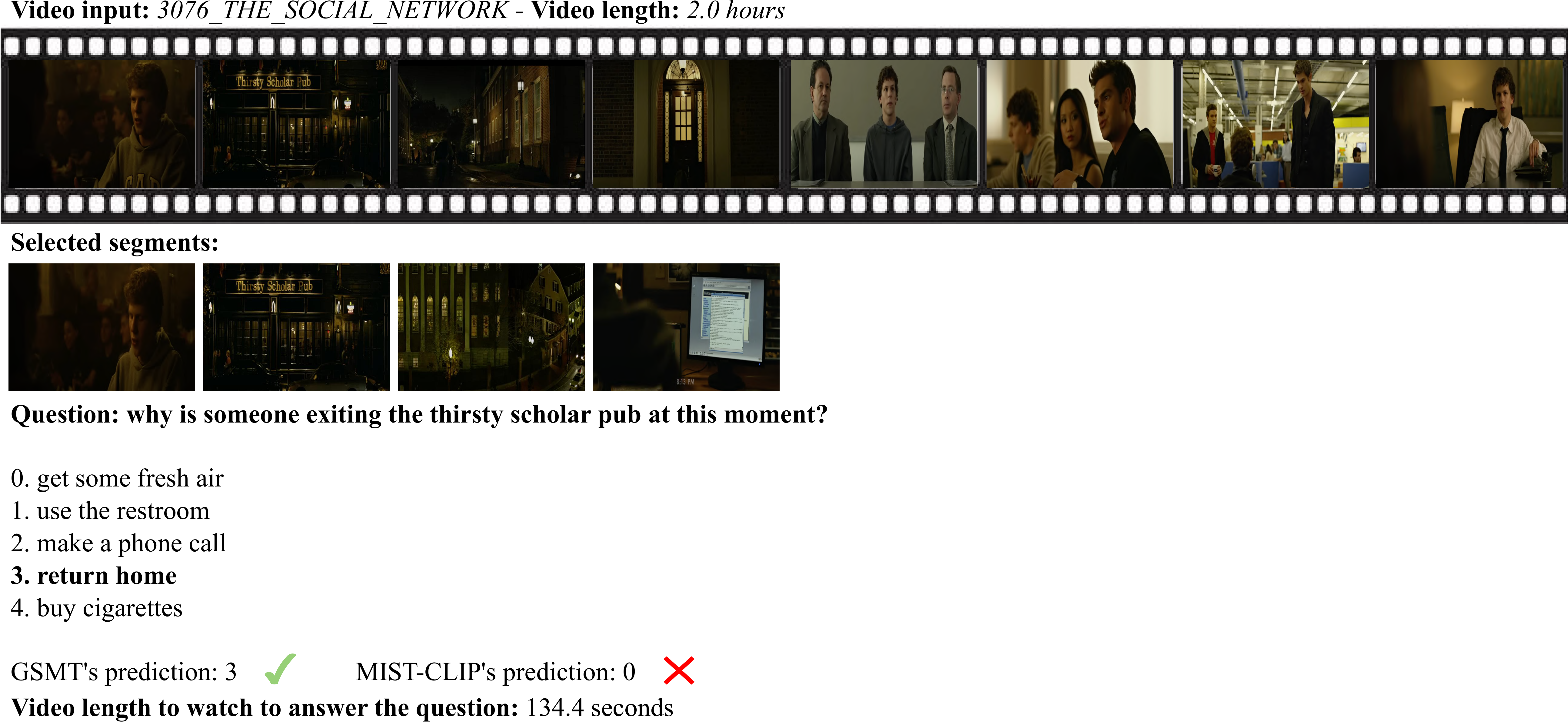}
\end{figure*}
\begin{figure*}[h!]
    \centering
    \includegraphics[width=0.9\linewidth]{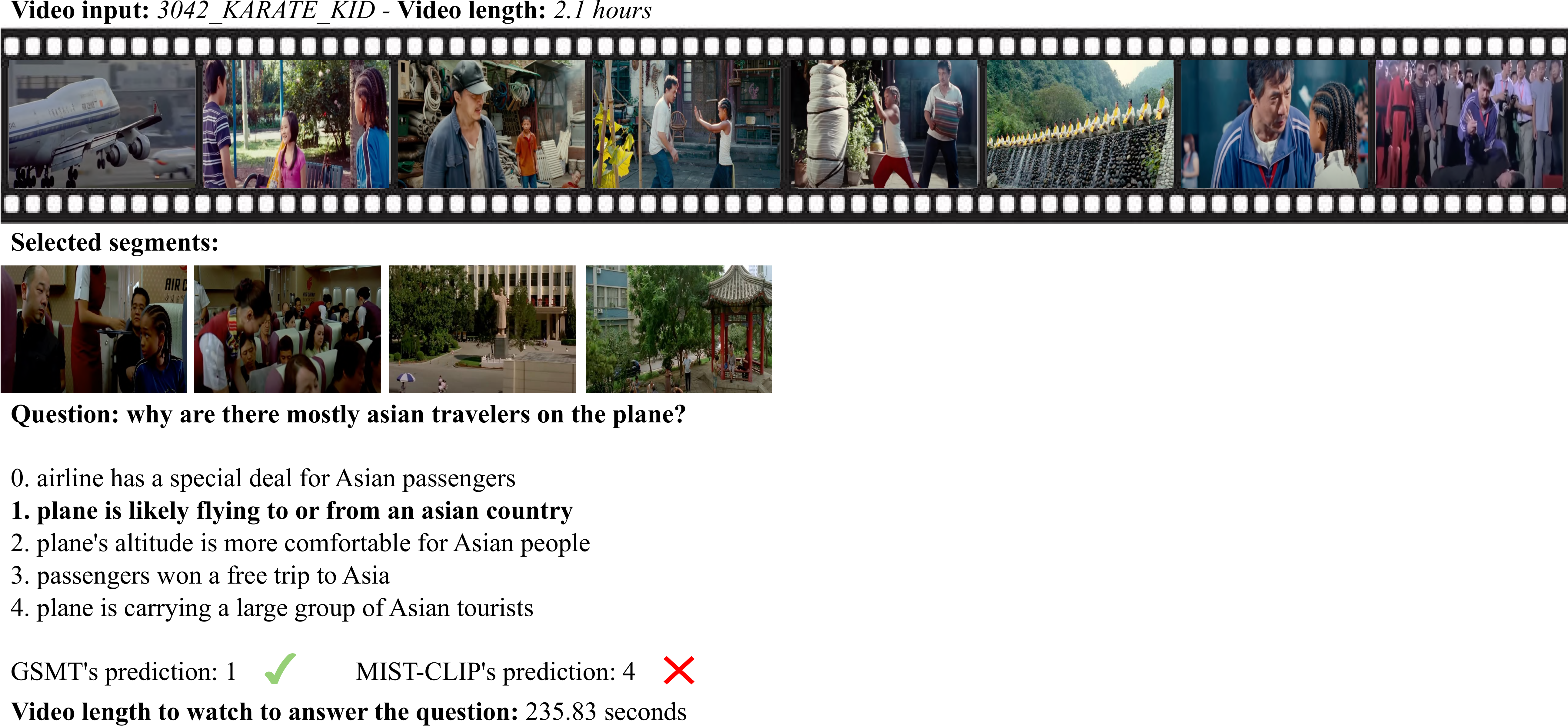}
\end{figure*}
\begin{figure*}[h!]
    \centering
    \includegraphics[width=0.9\linewidth]{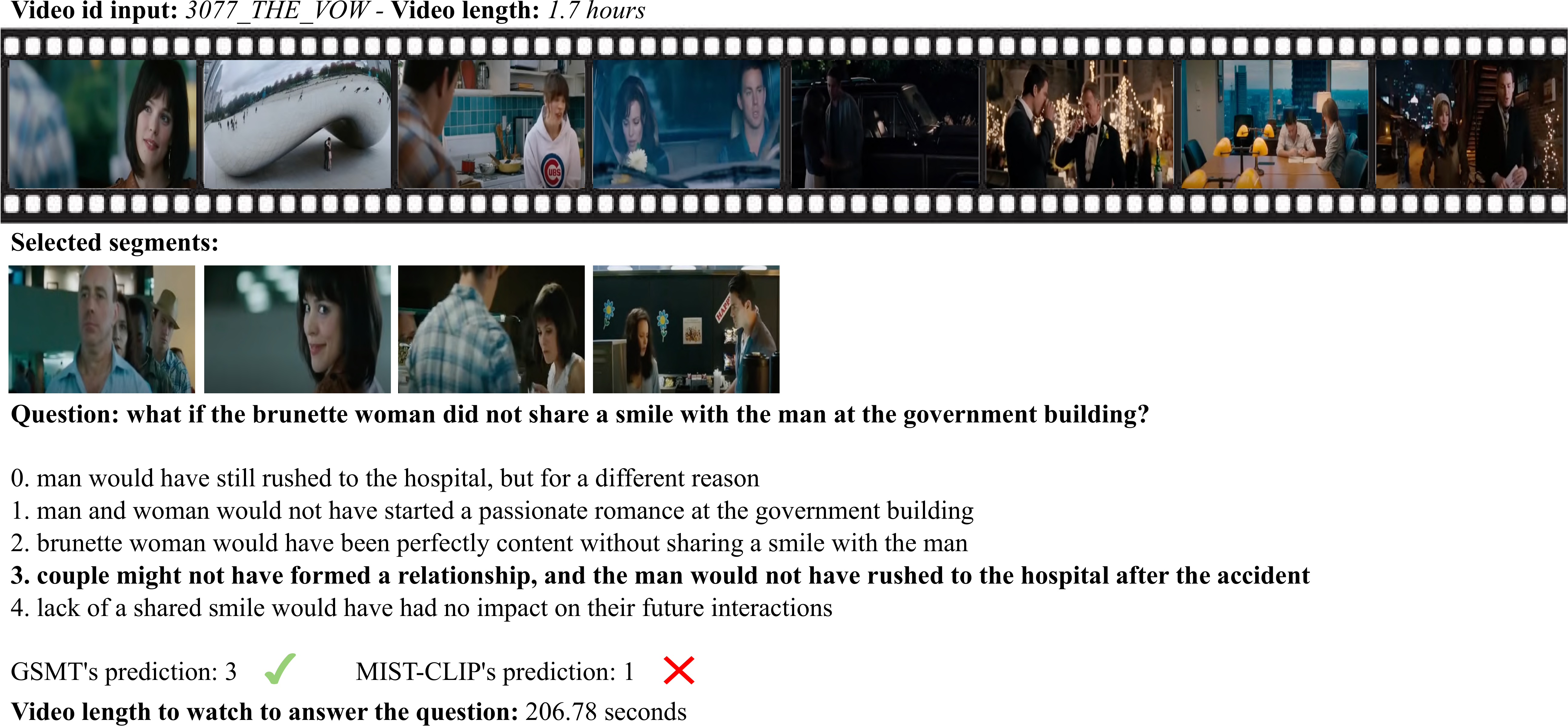}
\end{figure*}

\section{Dataset Statistics}
\label{app:dataset_statistics}
\begin{figure*}[h!]
    \centering
    \includegraphics[width=0.6\linewidth]{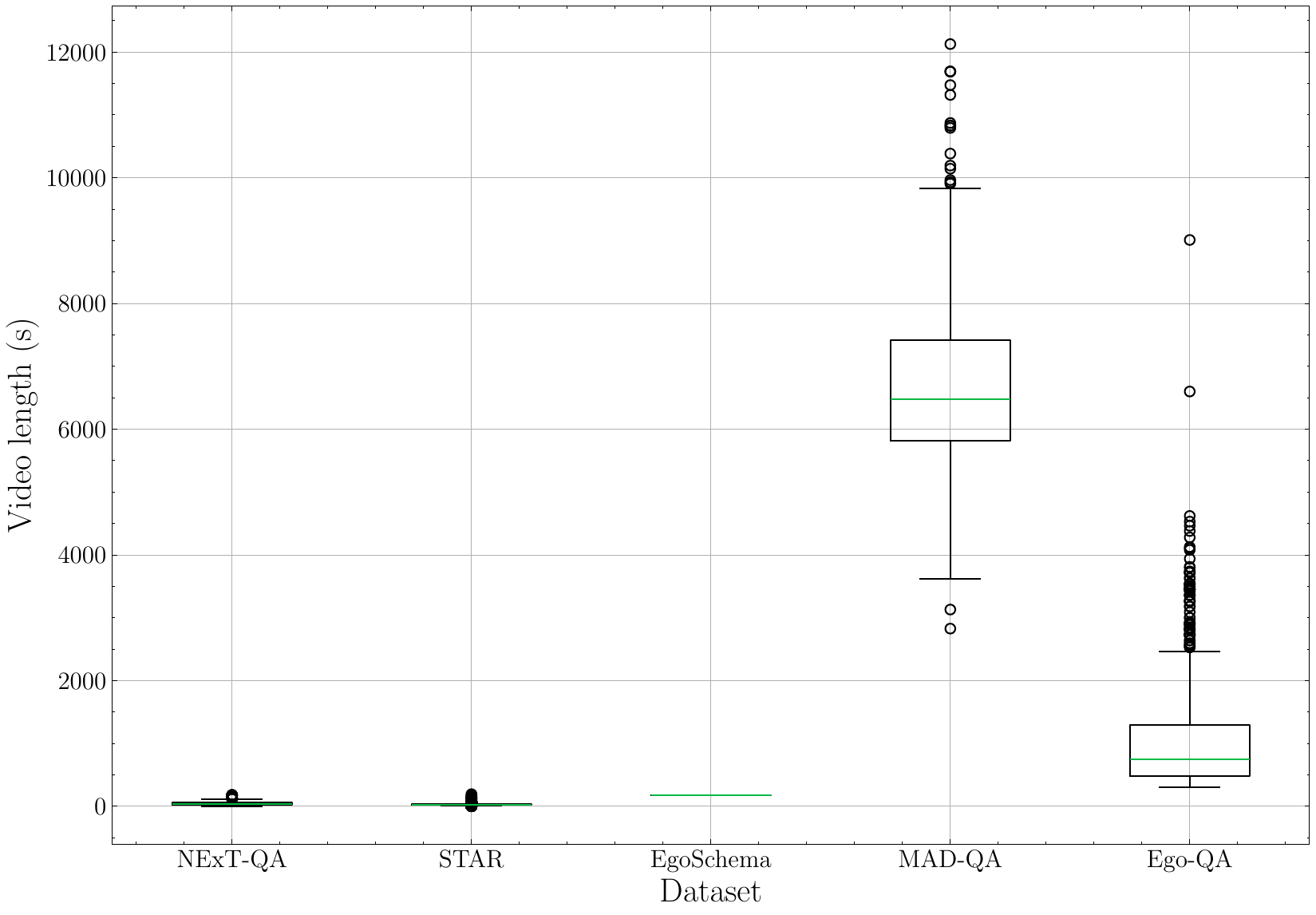}
    \caption{Distribution of video input duration of long-form videoQA datasets.}
    \label{fig:video_duration_distribution}
\end{figure*}

\begin{figure*}[h!]
    \centering
    \includegraphics[width=0.6\linewidth]{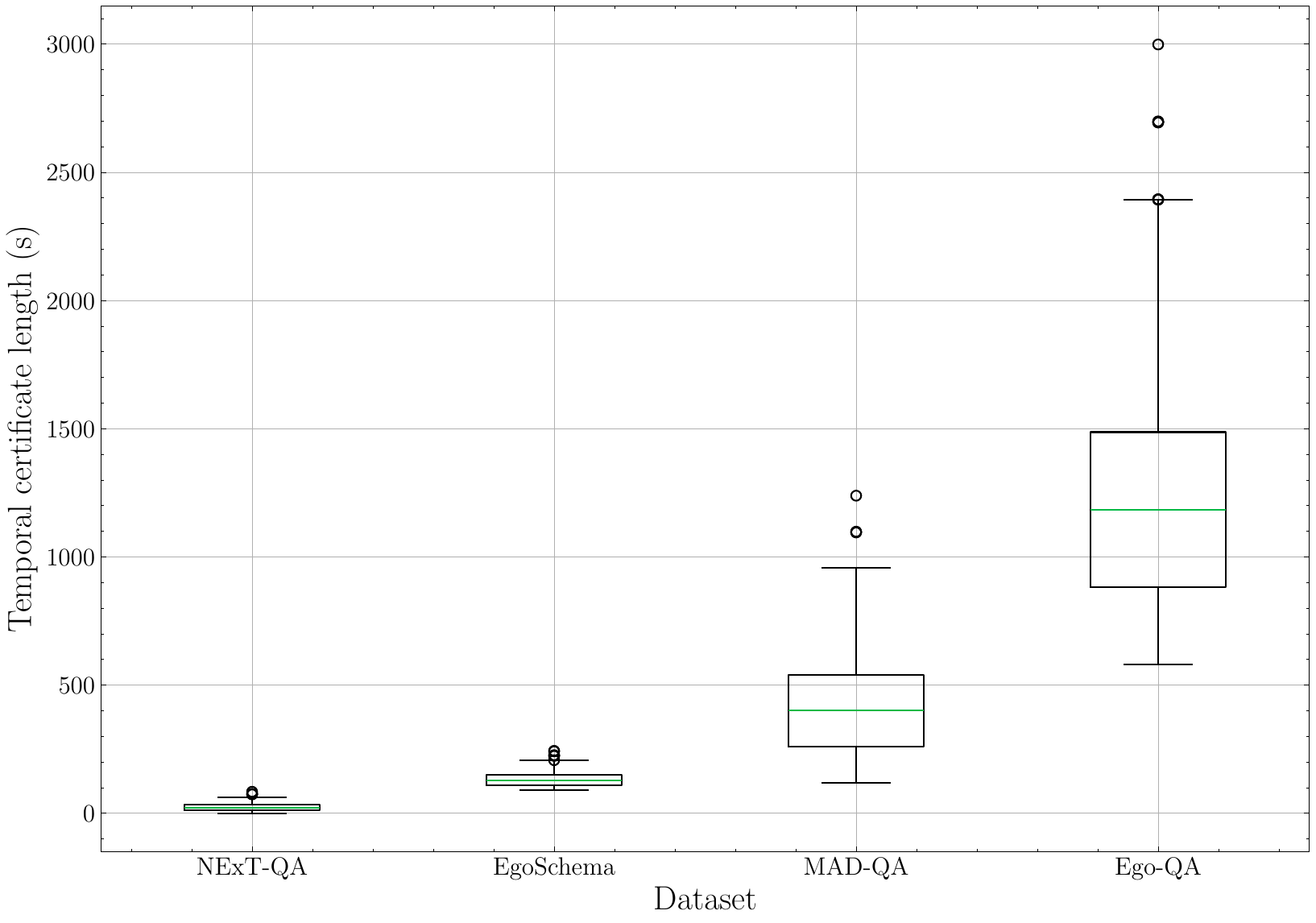}
    \caption{Distribution of temporal certificate lengths of long-form videoQA datasets.}
    \label{fig:temporal_certificate_distribution}
\end{figure*}

In this appendix, we provide the total number of questions and videos, and the average number of narration sentences per video in Table \ref{tab:statistics_egoqa_madqa}. We also visualize the distribution of video lengths of datasets widely used for long-form videoQA, \textit{i.e.} NExT-QA \citep{xiao2021next}, STAR \citep{wu2021star}, EgoSchema \citep{mangalam2023egoschema}, and our curated datasets MAD-QA and Ego-QA in Figure \ref{fig:video_duration_distribution}. In addition, we show the distribution of temporal certificate lengths, \textit{i.e.} video lengths humans need to watch to determine the answer, of standard long-form videoQA and our datasets in Figure \ref{fig:temporal_certificate_distribution}. As can be observed, our datasets exhibit longer video input length and also certificate length to verify the correct answer, validating their effectiveness to evaluate long-form videoQA performance.

\begin{table}[h!]
\centering
\resizebox{0.7\linewidth}{!}{
\begin{tabular}{l|c|c|c}
\hline
\multicolumn{1}{c|}{\textbf{Dataset}} & \textbf{\# Questions} & \textbf{\# Videos} & \textbf{\# Narration sentences per video (Avg)} \\ \hline
Ego-QA                               & 18838                        & 992                                           & 279.9                                                \\ 
MAD-QA                               & 15674                        & 650                                           & 641.3                                                \\ \hline
\end{tabular}}
\caption{Statistics of the curated Ego-QA and MAD-QA datasets.}
\label{tab:statistics_egoqa_madqa}
\end{table}

\end{document}